\documentclass[10pt, conference,a4paper]{IEEEtran}
\makeatletter
\long\def\@makecaption#1#2{\ifx\@captype\@IEEEtablestring%
\footnotesize\begin{center}{\normalfont\footnotesize #1}\\
{\normalfont\footnotesize\scshape #2}\end{center}%
\@IEEEtablecaptionsepspace
\else
\@IEEEfigurecaptionsepspace
\setbox\@tempboxa\hbox{\normalfont\footnotesize {#1.}~~ #2}%
\ifdim \wd\@tempboxa >\hsize%
\setbox\@tempboxa\hbox{\normalfont\footnotesize {#1.}~~ }%
\parbox[t]{\hsize}{\normalfont\footnotesize \noindent\unhbox\@tempboxa#2}%
\else
\hbox to\hsize{\normalfont\footnotesize\hfil\box\@tempboxa\hfil}\fi\fi}

\usepackage[compatibility=false]{caption}
\DeclareCaptionFont{quackfont}{\fontfamily{ptm}\fontsize{7.5pt}{9pt}\selectfont}
\usepackage[labelformat=simple,font=quackfont]{subcaption}

\makeatother
\usepackage{lipsum}
\makeatletter
\let\MYcaption\@makecaption
\usepackage{cite}
\usepackage{graphicx}
\usepackage{subcaption}
\usepackage{float}
\usepackage{refstyle}
\usepackage{multicol, blindtext}
\usepackage{amsmath,amssymb,amsfonts}
\usepackage{adjustbox}

\usepackage[font=scriptsize]{caption}
\usepackage[font=footnotesize]{caption}
\usepackage{stfloats}
\usepackage{lmodern,bm}                
\usepackage[T1]{sansmath}
\usepackage{xcolor}
\usepackage{algorithm, algorithmic}
\usepackage{enumitem}
\usepackage{dutchcal}

\begin{document}

\title{Skeleton-Split Framework using\\ 
Spatial Temporal Graph Convolutional Networks\\ 
for Action Recogntion}

\author{\IEEEauthorblockN{Motasem S. Alsawadi\IEEEauthorrefmark{1}\IEEEauthorrefmark{11} and Miguel Rio\IEEEauthorrefmark{1}}\\
\IEEEauthorblockA{\IEEEauthorrefmark{1}Department of Electronic and Electrical Engineering, University College London, WC1E 7JE UK}\\[-2.0ex]
\IEEEauthorblockA{\IEEEauthorrefmark{11}King Abdulaziz City for Science and Technology, Riyadh 12354, Saudi Arabia}\\[-2.0ex]
e-mail: (motasem.alsawadi.18; miguel.rio)@ucl.ac.uk, malswadi@kacst.edu.sa
}

\maketitle

\begin{abstract}
There has been a dramatic increase in the volume of videos and their related content uploaded to the internet. Accordingly, the need for efficient algorithms to analyse this vast amount of data has attracted significant research interest. An action recognition system based upon human body motions has been proven to interpret videos' contents accurately. This work aims to recognize activities of daily living using the ST-GCN model, providing a comparison between four different partitioning strategies: spatial configuration partitioning, full distance split, connection split, and index split.  To achieve this aim,  we present the first implementation of the ST-GCN framework upon the HMDB-51 dataset. We have achieved 48.88\% top-1  accuracy by using the connection split partitioning approach. Through experimental simulation, we show that our proposals have achieved the highest accuracy performance on the UCF-101 dataset using the ST-GCN framework than the state-of-the-art approach. Finally, accuracy of 73.25\% top-1 is achieved by using the index split partitioning strategy.
\end{abstract}


\IEEEpeerreviewmaketitle

\section{Introduction}\label{sec:introduction}
\IEEEPARstart{R}{ecently}, the amount of videos uploaded to the internet has increased substantially. According to Statista \cite{StatistaResearchDepartment}, by May 2019, more than 500 hours of video were uploaded to YouTube every minute, and the numbers did not slow down. Therefore, the need for robust algorithms to analyse this enormous amount of data has increased accordingly. \par
An action recognition system based upon human body motions is the most efficient way of interpreting videos' contents.  Several solutions have been proposed in this regard, and they vary from the analysis of optical flow  \cite{Bobick2001}, convolutional neural networks upon RGB images \cite{Tran2015} and more recently, the skeleton movements \cite{yan2018spatial}. The skeleton movements approach offers multiple advantages over the other solutions. The skeleton information is robust to changes in the illumination of the environment where the action takes place. Also, it is robust to changes in the background \cite{Keskes2021}. Moreover, the computational cost for training is considerably reduced for skeleton data consisting of only sets of joint cartesian coordinates. For these reasons, we have chosen this approach to define the premise of our proposed method.\par
\begin{figure}[!t]
    \centering
    \includegraphics[width=0.4\textwidth]{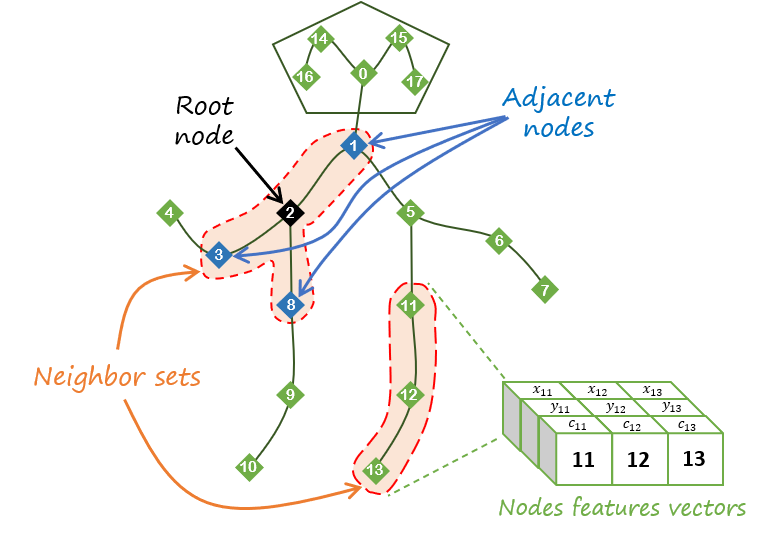}
    \caption{Skeleton components and the keypoints indexes of OpenPose layout.}
    \label{fig:keypoints}
\end{figure}

There are multiple sources to obtain the skeleton information from videos. Among these, the OpenPose library \cite{Cao2021a} is the simplest yet effective tool to accomplish this. This system receives a video clip as an input and outputs the 2D or 3D coordinates of the 18 main skeleton joints.  Each skeleton joint information consists of three values as (x, y, c), where x and y are the cartesian coordinates in the horizontal and vertical axis, respectively, and c represents the confidence score of the detected joint. The keypoints indexes of the OpenPose output layout are shown in Fig. \ref{fig:keypoints}.

\begin{figure*}[!t]

 \centering
     \begin{subfigure}[b]{0.45\textwidth}
         \centering
         \includegraphics[width=\textwidth]{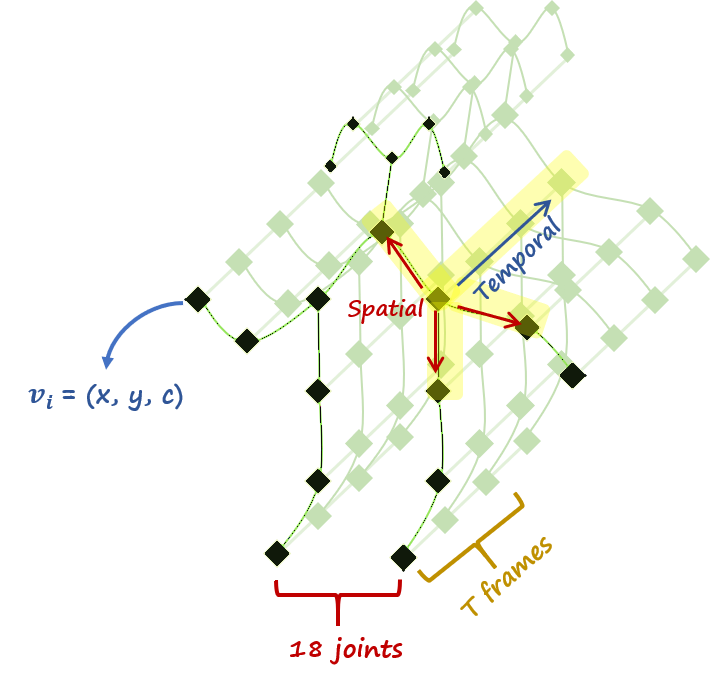}
         \caption{Spatial-temporal skeleton.}
         \label{fig:skeleton_rep}
     \end{subfigure}
     \hfill
     \begin{subfigure}[b]{0.45\textwidth}
         \centering
         \includegraphics[width=\textwidth]{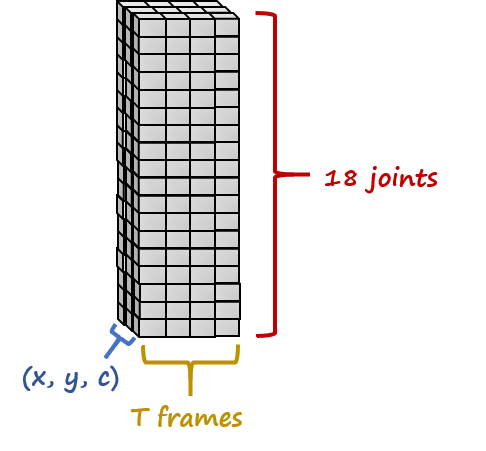}
         \caption{Tensor.}
         \label{fig:tensor_rep}
     \end{subfigure}
        \caption{Video clip representations.}
        \label{fig:video_rep}
\end{figure*}

Our study is based upon the proposal presented by Yan \textit{et al.} in \cite{yan2018spatial}. Instead of analysing the frames of a video by their pixel values (i.e., RGB images), the authors first represent the actors as a set of the main joints of the body using the OpenPose library \cite{Cao2021a}. Given the skeleton representation of the person performing the action, they model the skeleton joints as a set of vertices of a graph. On the other hand, the bones-like connections can be represented as the edges of the graph. Thus, the video clips are transformed from RGB image sequences to a sequence of skeleton joints. To achieve the action recognition, the authors proposed the Spatial-Temporal Graph Convolutional Neural Network (ST-GCN) model. As the name indicates, this framework can analyse both the spatial and the temporal relations between the set of nodes (i.e., the skeleton joints) during the performance of the action (Fig. \ref{fig:skeleton_rep}). Subsequently, the model is trained in an end-to-end manner using a Graph Convolutional Neural Network (GCN) architecture \cite{Zhou2018}.
Presently, there are multiple datasets available for research on human action recognition. Among these alternatives, the UCF-101 \cite{Soomro2012}, and the HMDB-51 \cite{Kuehne2011a} datasets are considered to be reference benchmarks.

\subsection{UCF-101}
The UCF-101 is the most commonly used benchmark human action dataset. Every video sample from this dataset is sourced from YouTube. The clip's duration varies from 1.06 sec to 71.04 sec and has a fixed frame rate of 25 fps and a fixed resolution of 320×240 pixels. This dataset provides a total of 13,320 clips classified into 101 action classes. These classes can be broadly divided into five major subsets: \textit{Human-Object Interaction}, \textit{Body-Motion Only}, \textit{Human-Human Interaction}, \textit{Playing Musical Instruments} and \textit{Sports} \cite{Soomro2012}.

\subsection{HMDB-51}
Similar to the UCF-101, the Human Motion Database (HMDB) is considered as one of the top 5 most popular datasets for action recognition \cite{Zhang2019b}. Aside from YouTube, the HMDB-51 dataset was collected from a wider range of sources (i.e., movies, Google videos, etc.).  For that reason, the height of all the samples was scaled to 240 pixels, and the width has was scaled to maintain the original video ratio. Furthermore, the frame rate was modified to have a fixed value of 30 fps. It provides a total of 6,766 video clips of 51 different classes. These classes can be broadly classified into 5 categories: \textit{General facial actions}, \textit{Facial actions with object manipulation}, \textit{General body movements}, \textit{Body movements with object interaction} and \textit{Body movements for human interaction} \cite{Kuehne2011a}. 

\subsection{ST-GCN additional layer: the M-Mask}
Complex movements can be inferred from a small set of representatives \emph{bright spots} on the joints of the human body \cite{Johansson1973}. However, not all the joints provide the same quality and quantity of information regarding the movement performed. Therefore, it is intuitive to assign a different level of importance to every joint in the skeleton. In the ST-GCN framework, the authors added a mask M (or M-mask) to each layer of the GCN to express the importance of each joint \cite{yan2018spatial}. This mask scales the contribution of each skeleton's joint according to the learned weights of the spatial graph network. According to their results, the proposed M-mask considerably improves their architecture's performance. Therefore, the authors constantly apply it to the ST-GCN network in their experiments.

\subsection{Our Contribution}
The convolution operation is not explicitly defined on graphs. Suppose a graph with no fixed structure (i.e., the quantity and arrangement of the nodes may vary), label mapping criteria need to be defined to perform the convolution process. For instance, the work of Yan \textit{et al.} in \cite{yan2018spatial} proposed three skeleton partition strategies (Uni-label, distance, and spatial configuration partitioning) to perform action recognition. This strategy was applied using ST-GCN upon the UCF-101 dataset \cite{Zheng2019}. However, to the best of our knowledge, there is no previous evidence of using the ST-GCN model on the HMDB-51 dataset for action recognition. In what follows, we summarise our contributions below: 
\begin{itemize}
    \item We present the first results of the ST-GCN model trained on the HMDB-51 dataset for action recognition. Moreover, we have used the previous skeleton extraction information of both the UCF-101 and the HMDB-51 datasets for the experiments.
    \item We have implemented our proposed partitioning strategies on the ST-GCN model \cite{Alsawadi2021} on the benchmark datasets (UCF-101 and the HMDB-51 datasets). 
    \item We provide a deep analysis of the impact of different batch sizes during training upon the accuracy performance of the output models using the both benchmark datasets.
    \item Additionally, we have provided the open-source skeleton information of the UCF-101 and HMDB-51 datasets for the research community\footnote{https://github.com/malswadi/skeleton\_ucf\_hmdb}.
\end{itemize}
The remainder of the paper is structured as follows: in \textbf{Section II}, we present the state-of-the-art skeleton-based systems that utilize the ST-GCN model for action recognition. In \textbf{Section III} we explain the constraints we have used in our experiments. The experimental results are described in depth in \textbf{Section IV}. Finally, \textbf{Section V} presents the summary and discussions.

\section{Action recognition using ST-GCN}
In order to perform the convolution operation, Yan \textit{et al.} \cite{yan2018spatial} first divided the skeleton into subsets of joints (\textit{i.e., neighbor-sets}). Each of these sets are composed by a \textit{root node} and its adjacent nodes (Fig. \ref{fig:keypoints}). On the other hand, each kernel has a size of $K$ x $K$.

In the same research \cite{yan2018spatial}, the authors used their architecture to recognize human actions upon the NTU-RGB+D \cite{Shahroudy2016} and the Kinetics \cite{Kay2017} dataset. They used the skeleton information of both datasets for their training. The NTU-RGB+D \cite{Shahroudy2016} provides the skeleton modality for their data with the main joints of the actors. However, the skeleton data is not available for the Kinetics dataset \cite{Kay2017}. Therefore, they initially extracted the skeleton data with the use of the OpenPose library \cite{Cao2021a} and released this data as the Kinetics-skeleton dataset \cite{yan2018spatial}. Once the skeleton information has been obtained, each video clip is modeled as a tensor (18,3, T), where T represents the length of the video as it is shown in Fig. \ref{fig:tensor_rep}. As a consequence, the data is prepared to perform the convolution process.

\subsection{Partitioning strategies}
For graphs with no specific order, priority criteria must be set in each neighbor-set to map each joint to a label. Hence, the convolution process can be performed, and network training can be possible. In \cite{yan2018spatial}, three neighbor set partitioning criteria were presented: Uni-labeling, Distance, and Spatial configuration partitioning. In the first approach, the kernel size K = 1. Therefore, all the joints in the neighbor set share the same label (label 0). In the second, the kernel size K = 2. The root node has the top priority (label 0), and the adjacent nodes share the same label (label 1). On the other hand, spatial configuration partitioning is more complex. 

\subsubsection{Spatial Configuration partitioning}
In this approach, the kernel size K = 3 and the center of gravity of the skeleton (average of the values on each joint axis across all the training set) are considered. Mathematically, the mapping for this strategy is defined with the following equation
\begin{equation}
l_{ti}(v_{tj}) = 
    \left\{ 
            \begin{array}{lcr}
             0  &   if  &   r_j=r_i\\
             1  &   if  &   r_j<r_i\\
             2  &   if  &   r_j>r_i
             \end{array}
   \right.
   ,
\label{eq:Strategy3}
\end{equation}
where $l_{ti}(v_{tj})$ represents the label mapping for the node $v_{tj}$, $r_j$ is the average value from the root node to the center of gravity and $r_i$ is the average value from the $i_{th}$ node to the center of gravity. Yan \textit{et al.} \cite{yan2018spatial} have reported a maximum performance accuracy on both of the NTU-RGB+D \cite{Shahroudy2016}, and the Kinetics-skeleton \cite{yan2018spatial} datasets using this partitioning strategy.\par

There have been multiple action recognition systems using the ST-GCN architecture \cite{Keskes2021, liutwo, Jiang2020, Yang2020}. Zheng \textit{et al.} \cite{Zheng2019} extracted the skeleton from the UCF-101 dataset in a similar manner as Yan \textit{et. al.} \cite{yan2018spatial} with the Kinetics \cite{Kay2017} and obtained 50.53\% top-1 accuracy using the spatial configuration partitioning for label mapping. Some additional hand-craft work needed to be done. They selected only the video clips on which the skeleton was detected during the first 250 frames.\par 

Recently, we were able to improve the ST-GCN performance upon the NTU-RGB+D \cite{Shahroudy2016} and the Kinetics \cite{Kay2017} benchmarks in \cite{Alsawadi2021}. As the base model \cite{yan2018spatial} proposed, we defined each neighbor set to contain a root node with its adjacent nodes. Nevertheless, in our previous work \cite{Alsawadi2021}, we considered a kernel size K=4. Thus, each of the nodes in the neighbor sets owns a separate label. The root node was set to have the highest priority in every split strategy (label 0). However, to define which of the adjacent nodes in the neighbor set has the highest priority in the label mapping, we have introduced three novel partitioning strategies: the \textit{full distance}, \textit{connection} and \textit{index splits}.

\subsubsection{Full distance split}
\begin{figure}[h]
    \centering
    \includegraphics[width=0.35\textwidth]{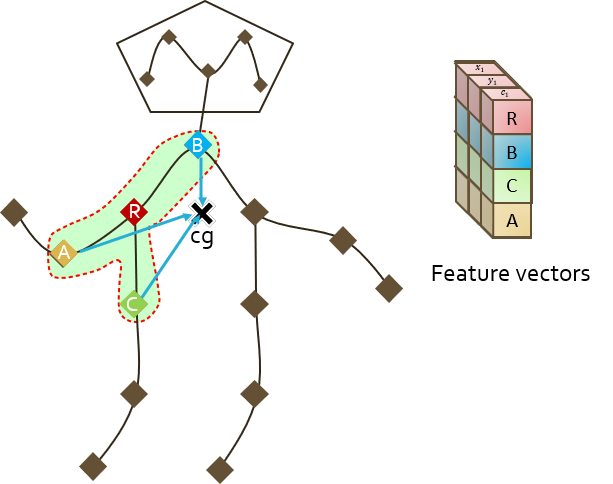}
    \caption{Full distance split.}
    \label{fig:full_distance}
\end{figure}

In this strategy, we took the contribution of the spatial configuration partitioning from Yan \textit{et al.} \cite{yan2018spatial} one step further. We considered the distance from \textit{every joint} in the neighbor set to the center of gravity of the skeleton. As it is shown in Fig. \ref{fig:full_distance}, the nearest the node is to the center of gravity, the highest priority it is assigned to it \cite{Alsawadi2021}. In the figure, the joint labeled as \textit{B} has the highest priority among the adjacent nodes because of its closeness to the center of gravity. To describe this strategy mathematically, a set $\mathcal{F}$ is defined. This set contains the Euclidean distances of the \textit{i}-th adjacent node $u_{ti}$ (of the root node $u_{tj}$) with respect to the center of gravity of the skeleton, sorted in ascending order as
\begin{equation}
     \mathcal{F}=\{f_{m|m=1,\cdots,N}\}
\end{equation}
where $N$ is the number of adjacent nodes to the root node $u_{tj}$. With this auxiliary set in place, the label mapping can be defined using the Eq. \ref{eq:full_distance}. 

\begin{equation}
     l_{ti}(u_{tj}) = 
    \left\{ 
            \begin{array}{lcr}
             0  &if& |u_{ti}-cg|_{2}=x_{r}\\
             m  &if& |u_{ti}-cg|_{2}=f_{m}
             \end{array}
   \right.
   ,
\label{eq:full_distance}
\end{equation}
where $l_{ti}$ represents the label map for each joint $u_{ti}$ in the neighbor set of the root node $u_{tj}$, $x_{r}$ is the Euclidean distance from the root node $u_{tj}$ to the center of gravity of the skeleton.

\subsubsection{Connection split}
\begin{figure}[h]
    \centering
    \includegraphics[width=0.35\textwidth]{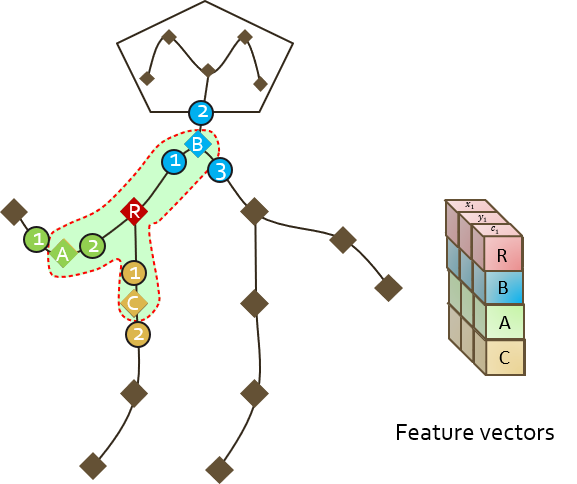}
    \caption{Connection split.}
    \label{fig:connection}
\end{figure}

For this partitioning criteria, the degree of each vertex (i.e., the joints) of the skeleton graph is considered. The higher degree, the higher priority \cite{Alsawadi2021}. For instance, consider the skeleton graph shown in Fig. \ref{fig:connection}. In the figure are indicated the connections of each of the adjacent nodes in the neighbor set. For this example, the root node has the top priority (label 0), and the node labeled as \textit{B} has the next priority (label 1). Given that both nodes \textit{A} and \textit{C} have the same degree, we considered them with the same priority; hence, their priority is set randomly.

To define the label mapping, we describe a set $\mathcal{C}$ as the degree values of each of the $N$ adjacent nodes of the root node sorted in descending order as follows
\begin{equation}
    \mathcal{C}=\{c_{m|m=1,\cdots,N}\}
\end{equation}

Given the set $\mathcal{C}$ defined, the label mapping can be obtained using Eq. \ref{eq:index_eq}. 

\begin{equation}
     l_{ti}(u_{tj}) = 
    \left\{ 
            \begin{array}{lcr}
                0 &if&d(u_{ti})=d_{r}\\
                m &if&d(u_{ti})=d_{m}\\
             \end{array}
   \right.
   ,
\label{eq:index_eq}
\end{equation}
where $l_{ti}$ represents the label map for each joint $u_{ti}$ in the neighbor set of the root node $u_{tj}$ and $d_{r}$ is the degree corresponding the root node. 

\subsubsection{Index split}
\begin{figure}[h]
    \centering
    \includegraphics[width=0.35\textwidth]{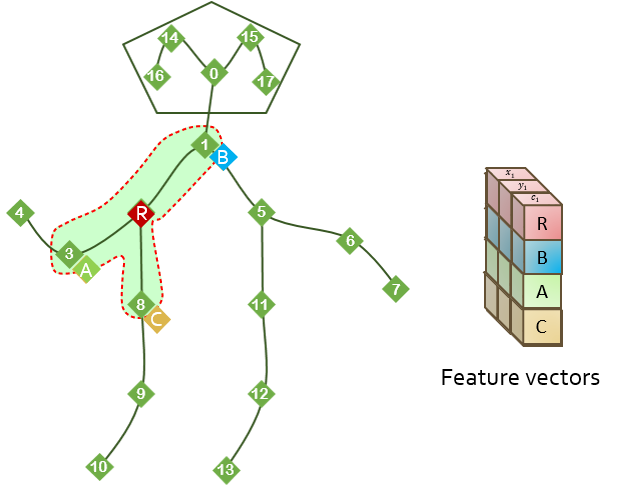}
    \caption{Index split.}
    \label{fig:index}
\end{figure}

For this strategy, we considered the OpenPose \cite{Cao2021a} output keypoints shown in Fig. \ref{fig:keypoints}. The priority criteria are defined as follows: the smallest value of the keypoint index, the highest priority \cite{Alsawadi2021}. For instance, consider the neighbor set shown in Fig. \ref{fig:index}. Like the other partition strategies, the highest priority is assigned to the root node (label 0). Subsequently, the adjacent with the highest priority is given to the node labeled as \textit{B} because its keypoint index is the smallest (index 1). Finally, the node labeled as \textit{A} and the node labeled as \textit{C} have the second and third priority, respectively.

Similar to the other split strategy, we defined an auxiliary set $\mathcal{P}$ with the keypoint index values of the adjacent nodes. 

\begin{equation}
    \mathcal{P}=\{p_{m|m=1,\cdots,N}\}
\end{equation}

The values of $\mathcal{P}$ are ascendant ordered. Then, the label mapping is obtained using Eq. \ref{eq:connection_eq}.
\begin{equation}
     l_{ti}(u_{tj}) = 
    \left\{ 
            \begin{array}{lcr}
                0 &if&ind(u_{ti})=in_{r}\\
                m &if&ind(u_{ti})=p_m\\
             \end{array}
   \right.
   ,
\label{eq:connection_eq}
\end{equation}
where $l_{ti}$ and $ind(u_{ti})$ represent the label map and the index keypoint value of the $i_{th}$ joint, respectively; and $in_{r}$ is the index of the keypoint corresponding to the root node $u_{tj}$.

\section{Experimental settings}
Given that the skeleton representation of the actors is not provided for either the UCF-101 \cite{Soomro2012} or the HMDB-51 \cite{Kuehne2011a}, we first extract that skeleton representation from both datasets. Similarly to \cite{yan2018spatial}, we used the Open-Pose library to extract the skeleton data for our evaluation. The library installation was oriented to be compatible with the Ubuntu 18.04 environment.\par 

We followed the experiment guidelines provided in \cite{yan2018spatial}. First, the resolution of each video sample has been resized into a fixed dimension of 340 × 256 pixels. Second, the set of resized image frames of the video samples is input to the Open-Pose algorithm. Third, due to the variability of the duration of each clip, a fixed duration of 300 frames has been proposed. Therefore, if any video clip has less than 300 frames, we repeat the initial frames until we reach the amount needed. Otherwise, if the video clip exceeds the frame number, we trim it. Consequently, the Spatio-temporal information of the skeleton of each video sample can be represented as a tensor with shape (18, 3, 300). By setting the \textit{T} value to 300, our output tensor is illustrated in Fig. \ref{fig:tensor_rep}. In the fourth step, we considered the joint recognition score provided by the Open-Pose algorithm (i.e., the \textit{C} value). After several experiments, we concluded to consider for training only those videos with more than 50\% skeleton joint recognition.\par
Additionally, we only considered those samples with a maximum of two people performing the action. Finally, the Spatio-temporal skeleton data of the UCF-101 and the HMDB-51 video that fulfilled those quality criteria is collected during the \textit{Data extraction} stage. We have iterated through the video clips of the datasets and saved the skeleton information as JSON files. An independent JSON file has been exported for each video sample. Thus, the outcome of this process are 13,320 and 6,766 JSON files with the skeleton information of the UCF-101 and the HMDB-51 datasets, respectively. These files are publicly available for the research community. \footnotemark[\value{footnote}] 

\subsection{Training Details}
We utilized the PyTorch framework \cite{Paszke2017} for deep learning modelling to execute our experiments. The experiment process is composed in 3 stages: \textit{Data Splitting}, \textit{ST-GCN Model Setup} and the \textit{Model Training}. The first stage divides each of the datasets mentioned above into two subsets: the training and the validation sets. For our experiments, we considered a 3:1 relation for training and validation split, respectively. Then, the second stage aims to prepare the ST-GCN architecture to be trained using the spatial configuration partitioning strategy proposed by Yan \textit{et al.} \cite{yan2018spatial} and also with the use of the our previously proposed split processes presented in \cite{Alsawadi2021}. \par
Finally, in the \textit{Model Training} stage, we performed the experiments of the implementation of the ST-GCN model using the spatial configuration partitioning \cite{yan2018spatial}. During this stage, we utilized the enhanced split strategies proposed in \cite{Alsawadi2021} in our experiments to find the partitioning approach that offered the best performance in terms of accuracy. To provide a valid comparison, we included the M-mask layer in the architecture during experimentation. Additionally, we perform a further analysis without the M-mask implementation. 

Every model has been trained using the stochastic gradient descent (SGD) with learning rate decay as optimization algorithm. Also, all the models training started with a learning rate value of 0.1. The models were trained for 80 epochs and the learning rate decays by a factor of 0.1 every \(10^{th}\) epoch, starting from the epoch number 20. Additionally, in order to avoid overfitting on the datasets, a weight decay value of 0.0001 has been considered.\par
One experiment setting criteria was to find the optimal batch size. This hyperparameter allows the model to adjust its parameters during optimization with respect to a small subset of training samples called \textit{mini batches} \cite{Goodfellow-et-al-2016}. The optimization algorithm requires a lower computational cost to update the weights by training the network in mini-batches. If the batch size is too small, the learned parameters in each step of the gradient descent tend to be less robust, given that the weights were updated from a set of samples with minor variation; if the batch size is too big, the computational cost increases accordingly. Therefore, we proposed this hyper-parameter to be one of the experiment's definition criteria. We performed the experiments using different batch sizes values. These vary from 8, 16, 32, 64, and 128.

\section{Results}
The experiment's outcome for each benchmark dataset is presented separately in different sections. The results correspond to the models with the best performance in terms of accuracy. The accuracy values shown were obtained using top-1 criteria. 

\subsection{UCF-101}
As mentioned in the previous section, we vary the implemented partition strategy in the ST-GCN architecture. Additionally, we performed experiments with and without the implementation of the M-mask. In Table \ref{table:batch_ucf_performance} are shown the results of these experiments upon the UCF-101. The "Y" ("Yes") and "N" ("No") values in the "M-mask" column represent whether the M-mask layer was implemented or not in that experiment, respectively. It can be noticed that, in most experiments, the output model tends to be more robust as the batch size increases. 

\begin{table}[ht]
\caption{Experiments performance upon UCF-101}
\centering
\begin{tabular}{c c c c}
\hline\hline 
 \centering Method & Batch size & M-mask & Accuracy \\ [0.5ex] 
\hline 
\centering Spatial C.P.   
& 8 & Y & 46.42\%  \\
&   & N & 65.36\%  \\
& 16 & Y &  68.71\%  \\
&   & N & 65.89\%  \\
& 32 & Y &  68.96\%  \\
&   & N & 68.55\%  \\
& 64 & Y &  70.47\%  \\
&   & N & 68.18\%  \\
& \textbf{128}  & \textbf{N} & \textbf{70.72\%}  
\\[1ex] 
\hline

\centering Full Distance Split 
& 8 & Y &  48.51\%  \\
&   & N & 58.73\%  \\
& 16 & Y &  61.02\%  \\
&   & N & 67.89\%  \\
& 32 & Y &  69.16\%  \\
&   & N & 66.30\%  \\
& \textbf{64} & \textbf{Y} &  \textbf{70.43}\%  \\
&   & N & 68.59\%  \\
& 128  & N & 66.91\%  
\\[1ex] 
\hline

\centering Connection Split  
& 8 & Y &  63.03\%  \\
&   & N & 63.48\%  \\
& 16 & Y &  64.46\%  \\
&   & N & 62.99\%  \\
& 32 & Y &  70.88\%  \\
&   & N & 69.41\%  \\
& \textbf{64} & \textbf{Y} &  \textbf{70.96\%}  \\
&   & N & 68.18\%  \\
& 128 & N &  70.35\%  
\\[1ex] 
\hline

\centering Index Split  
& 8 & Y &  56.61\%  \\
&   & N & 58.24\%  \\
& 16 & Y &  69.33\%  \\
&   & N & 62.70\%  \\
& 32 & Y &  68.34\%  \\
&   & N & 68.34\%  \\
& 64 & Y &  72.31\%  \\
&   & N & 72.19\%  \\
& \textbf{128} & \textbf{N} & \textbf{73.25\%}  \\
[1ex] 
\hline 
\end{tabular}
\label{table:batch_ucf_performance}
\end{table}

By analysing the output values of the experiments shown in Table \ref{table:batch_ucf_performance}, we have created Table \ref{table:m_mask_ucf_performance} with the results with the M-mask layer proposed by Yan \textit{et al.} \cite{yan2018spatial}. To provide a comparative, we have also included the outcome of the previous ST-GCN implementation performed by Zheng \textit{et al.} \cite{Zheng2019} in this table.

\begin{table}[ht]
\caption{UCF-101 performance using M-mask}
\centering
\begin{tabular}{c p{2.7cm} c}
\hline\hline 
 & \centering Method & Accuracy \\ [0.5ex] 
\hline 
ST-GCN & \centering Spatial Configuration Partitioning & 70.47\%  \\
Zheng \textit{et al.} \cite{Zheng2019} & \centering Spatial Configuration Partitioning  & 50.53\% \\
Alsawadi and Rio \cite{Alsawadi2021} & \centering Full Distance Split  & 70.43\% \\
Alsawadi and Rio \cite{Alsawadi2021} & \centering Connection Split & 70.96\% \\
\textbf{Alsawadi and Rio \cite{Alsawadi2021}} & \centering \textbf{Index Split} & \textbf{72.31}\% \\ [1ex] 
\hline 
\end{tabular}
\label{table:m_mask_ucf_performance}
\end{table}

The model with M-mask implementation that achieved the best accuracy performance was trained using a 64 batch size and utilized the index split partitioning strategy. It has achieved 1.84\% of accuracy improvement with respect to the spatial configuration partitioning approach proposed by Yan \textit{et al.} in \cite{yan2018spatial}. Moreover, this model enhances the previous state-of-the-art results by 21.78\%.

\begin{table}[ht]
\caption{UCF-101 performance without M-mask}
\centering
\begin{tabular}{c p{2.7cm} c}
\hline\hline 
 & \centering Method & Accuracy \\ [0.5ex] 
\hline 
ST-GCN & \centering Spatial Configuration Partitioning & 70.72\%  \\
Alsawadi and Rio \cite{Alsawadi2021} & \centering Full Distance Split  & 68.59\% \\
Alsawadi and Rio \cite{Alsawadi2021} & \centering Connection Split     & 70.35\% \\
\textbf{Alsawadi and Rio \cite{Alsawadi2021}} & \centering \textbf{Index Split}          & \textbf{73.25\%} \\ [1ex] 
\hline 
\end{tabular}
\label{table:no_m_mask_ucf_performance}
\end{table}

On the other hand, Table \ref{table:no_m_mask_ucf_performance}
shows that the accuracy performance increased when the M-mask implementation is not considered in the ST-GCN architecture. Again, the index split partitioning strategy allowed the ST-GCN model to achieve the best accuracy performance. For this model, a batch size of 128 was considered. This solution enhanced the spatial configuration partitioning model approach proposed by Yan \textit{et al.} in \cite{yan2018spatial} by 2.53\%.\par
\begin{figure}[h]
    \centering
    \includegraphics[width=0.45\textwidth]{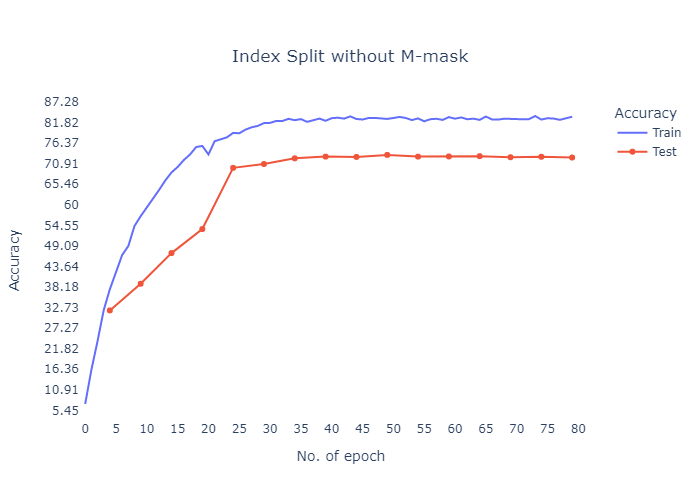}
    \caption{Best UCF-101 Model Training Process.}
    \label{fig:ucf_index}
\end{figure}

Therefore, we can reach the highest accuracy performance by using the index split partitioning approach upon the ST-GCN architecture without the M-mask implementation. We have evaluated the model for each of the five epochs. The outcome of this model during training is shown in Fig. \ref{fig:ucf_index}. The evaluation of the training and the test set are shown in red and blue curves, respectively. 

\subsection{HMDB-51}
The results corresponding to the different experiments upon the HMDB-51 dataset are shown in Table \ref{table:batch_hmdb_performance}. Similar to the outcome obtained in Table \ref{table:batch_ucf_performance}, in most of the experiments, the accuracy performance tends to improve as the batch size increases.

\begin{table}[ht]
\caption{Experiments performance upon HMDB-51}
\centering
\begin{tabular}{c c c c}
\hline\hline 
 \centering Method & Batch size & M-mask & Accuracy \\ [0.5ex] 
\hline 
\centering Spatial C.P.   
& 8 & Y & 37.34\%  \\
&   & N & 40.77\%  \\
& 16 & Y &  44.39\%  \\
&   & N & 41.08\%  \\
& 32 & Y &  43.89\%  \\
&   & N & 45.45\%  \\
& \textbf{64} & Y &  45.64\%  \\
&   & \textbf{N} & \textbf{46.82\%}  \\
& 128 & N & 44.64\% \\[1ex] 
\hline

\centering Full Distance Split 
& 8 & Y &  41.77\%  \\
&   & N & 33.23\%  \\
& 16 & Y &  38.97\%  \\
&   & N & 42.08\%  \\
& 32 & Y &  33.23\%  \\
&   & N & 45.51\%  \\
& \textbf{64} & Y &  42.02\%  \\
&   & \textbf{N} & \textbf{45.82\%}  \\
& 128 & N & 45.26\%  \\[1ex] 
\hline

\centering Connection Split  
& 8 & Y &  23.63\%  \\
&   & N & 39.34\%  \\
& 16 & Y &  43.27\%  \\
&   & N & 40.84\%  \\
& 32 & Y &  40.52\%  \\
&   & N & 41.52\%  \\
& 64 & Y &  32.29\%  \\
&   & N & 47.19\%  \\
& \textbf{128} & \textbf{N} &  \textbf{48.88\%}  \\[1ex] 
\hline

\centering Index Split  
& 8 & Y &  38.97\%  \\
&   & N & 34.91\%  \\
& 16 & Y &  35.47\%  \\
&   & N & 46.57\%  \\
& 32 & Y &  43.20\%  \\
&   & N & 45.51\%  \\
& \textbf{64} & \textbf{Y} &  \textbf{47.69\%}  \\
&   & N & 43.39\%  \\
& 128 & N &  46.51\%  \\
[1ex] 
\hline 
\end{tabular}
\label{table:batch_hmdb_performance}
\end{table}

 There is no previous application of the ST-GCN model upon the HMDB-51 dataset to the author's knowledge. Hence, the table only contains the results of the present study using the different partitioning strategies.

\begin{table}[ht]
\caption{HMDB-51 performance using M-mask}
\centering
\begin{tabular}{c  p{2.7cm}  c}
\hline\hline 
 & \centering Method & Accuracy \\ [0.5ex] 
\hline 
ST-GCN & \centering Spatial Configuration Partitioning & 45.64\%  \\
Alsawadi and Rio \cite{Alsawadi2021} & \centering Full Distance Split  & 42.02\% \\
Alsawadi and Rio \cite{Alsawadi2021} & \centering Connection Split & 45.89\% \\
\textbf{Alsawadi and Rio \cite{Alsawadi2021}} & \centering \textbf{Index Split} & \textbf{47.69\%} \\ [1ex] 
\hline 
\end{tabular}
\label{table:m_mask_hmdb_performance}
\end{table}

Table \ref{table:m_mask_hmdb_performance} contains the highest performance achieved with each partitioning strategy with M-mask implementation upon the HMDB-51 dataset. As indicated in bold letters, the highest value was performed using the index split partition strategy. This model was trained by choosing a training batch size of 64. It has reached more than 2\% accuracy improvement with respect to the spatial configuration partitioning proposed by Yan \textit{et al.} in \cite{yan2018spatial}. Additionally, it can be noticed that also the connection split outperformed the spatial configuration partitioning outcome.

\begin{table}[ht]
\caption{HMDB-51 performance without M-mask}
\centering
\begin{tabular}{c p{2.7cm} c}
\hline\hline 
 &\centering Method & Accuracy \\ [0.5ex] 
\hline 
ST-GCN & \centering Spatial Configuration Partitioning & 46.82\%  \\
Alsawadi and Rio \cite{Alsawadi2021} & \centering Full Distance Split  & 45.82\% \\
\textbf{Alsawadi and Rio \cite{Alsawadi2021}} & \centering \textbf{Connection Split} & \textbf{48.88\%} \\
Alsawadi and Rio \cite{Alsawadi2021} & \centering Index Split          & 46.57\% \\ [1ex] 
\hline 
\end{tabular}
\label{table:no_m_mask_hmdb_performance}
\end{table}

\begin{figure}[h]
    \centering
    \includegraphics[width=0.45\textwidth]{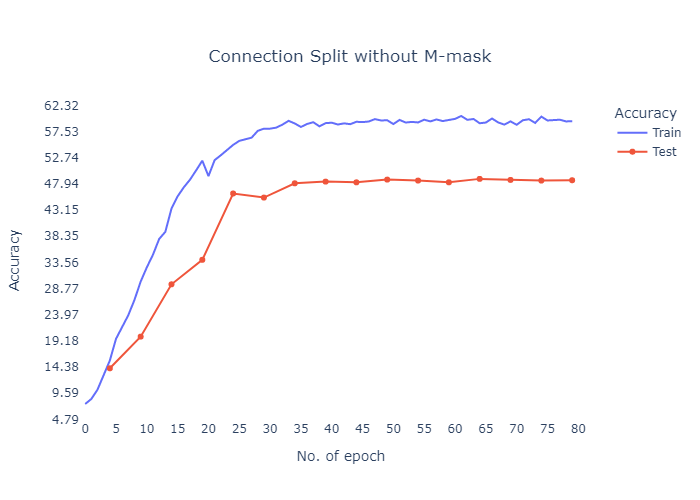}
    \caption{Best HMDB-51 Model Training Process.}
    \label{fig:hmdb_connection}
\end{figure}

In the experiments with no M-mask implementation shown in Table \ref{table:no_m_mask_hmdb_performance}. The partitioning strategy that achieved the highest accuracy performance was the connection split. This model was trained using a batch size of 128. The result obtained with this model outperformed with more than 2\% the outcome of the ST-GCN architecture without M-mask implementation using the spatial configuration partitioning. The training process performance of this model is shown in Fig. \ref{fig:hmdb_connection}. As Fig. \ref{fig:ucf_index}, the evaluation upon the training and the test set is shown in the colours red and blue, respectively. We have tested the performance of the trained model for every five epochs.  

\section{Conclusion}
In this paper, we have proposed novel action recognition method using ST-GCN model by exploiting partitioning strategies: \textit{spatial configuration paritioning}, \textit{full distance split}, \textit{connection split} and \textit{index split}. We have presented the first implementation of the ST-GCN framework on the HMDB-51 \cite{Kuehne2011a} dataset achieving 48.88\% top-1 accuracy by using the connection split partitioning approach. Our proposal outperforms the state-of-the-art using the ST-GCN framework on the UCF-101. Our results further show performance superiority over the most recent related work proposed in \cite{Alsawadi2021} with much lower training and computational inference costs and structural simplicity.\par 

The difference in the amount of training data impacted considerably in the final performance. The UCF-101 provides more than twice the amount of samples for training than the HMDB-51 counterpart. Therefore, the learning achieved with this dataset is more robust to changes in the input data than the model obtained with the second set. This is clearly demonstrated in the accuracy result values of Tables \ref{table:batch_ucf_performance} and \ref{table:batch_hmdb_performance}.

As future work, we propose increasing the size of nodes in the neighbor sets to capture the relationships between joints that are distant from each other. We believe that the more details we can capture of each movement, the more we can model the action to increase the overall accuracy. 

\bibliographystyle{IEEEtran}
\bibliography{ref}

\end{document}